\documentclass[10pt,twocolumn,letterpaper]{article}

\newcommand{\comment}[1]{}

\newcommand{\etal}{\emph{et al. }}

\newcommand{\bp}{\mathbf{p}}

\newcommand{\bu}{\mathbf{u}}

\newcommand{\bB}{\mathbf{B}}

\newcommand{\bI}{\mathbf{I}}

\newcommand{\bX}{\mathbf{X}}
\newcommand{\bY}{\mathbf{Y}}

\usepackage{cvpr}
\usepackage{times}
\usepackage{epsfig}
\usepackage{amsmath}
\usepackage{amssymb}
\usepackage{caption}
\usepackage{xcolor}
\usepackage{colortbl}
\usepackage{balance}

\usepackage[breaklinks=true,bookmarks=false]{hyperref}

\cvprfinalcopy 

\def\cvprPaperID{3916} 

\ifcvprfinal\fi
\begin{document}

\title{Unsupervised Person Image Synthesis in Arbitrary Poses}

\author{Albert Pumarola\\
\and
Antonio  Agudo\\
\and
Alberto  Sanfeliu\\
\and 
Francesc  Moreno-Noguer\\
\and
Institut de Rob\`otica i Inform\`atica Industrial (CSIC-UPC)\\ 08028, Barcelona, Spain\\
}

\twocolumn[{%
	\renewcommand\twocolumn[1][]{#1}%
\maketitle

\begin{center}
	\centering
		\includegraphics[width=1\textwidth,height=5.0cm]{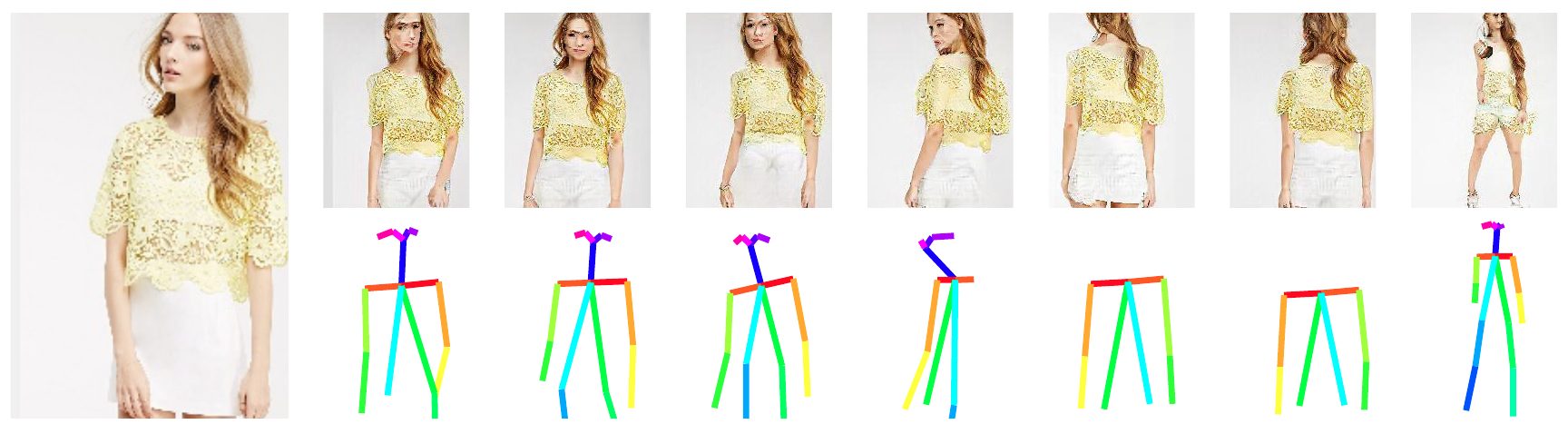}
\vspace{-8mm}			
\captionof{figure}{Given an original image of a person (left) and a desired body pose defined by a 2D skeleton (bottom-row), our model generates new photo-realistic images of the person under that pose (top-row).   The main contribution of our work is to train this generative model with unlabeled data.}
\label{fig:intro}
\end{center}%
}]

\begin{abstract}
\vspace{-0.2cm}
We present a novel approach for synthesizing photo-realistic images of people in arbitrary poses using generative adversarial learning. Given an input image of a person and a desired pose represented by a 2D skeleton, our model renders the image of the same person under the new pose, synthesizing novel views of the parts visible in the input image and hallucinating those that are not seen. This problem has recently been addressed in a supervised manner~\cite{ma2017pose,zhao2017multiview}, i.e., during training the ground truth images under the new poses are given  to the network. We go beyond these approaches by proposing a fully unsupervised strategy. We tackle this challenging scenario by splitting the problem into two principal subtasks.  First, we consider a pose conditioned bidirectional generator that maps back the initially rendered image to the original pose, hence being directly comparable to the input image without the need to resort to any training image. Second, we devise a novel loss function that incorporates content and style terms, and aims at producing images of high perceptual quality. Extensive experiments conducted on the DeepFashion dataset demonstrate that the images rendered by our model are very close in appearance to those obtained by fully supervised approaches.

\end{abstract}

\section{Introduction}
Being able to generate novel photo-realistic views of a person  in an arbitrary pose from a single image would open the door to many new exciting applications in  different areas, including   fashion and e-commerce business, photography technologies to automatically edit and animate still images, and the movie industry to name a few. Addressing this task without explicitly capturing the underlying processes involved in the image formation such as estimating the 3D geometry of the body, hair and clothes, and the appearance and reflectance models of the visible and occluded parts seems an extremely complex endeavor. Nevertheless, Generative Adversarial Networks (GANs) ~\cite{goodfellow2014generative} have shown impressive results in rendering new realistic images, e.g., faces~\cite{karras2017progressive,radford2015unsupervised}, indoor scenes~\cite{wang2016generative} and clothes~\cite{iccv2017fashiongan}, by directly learning a generative model  from data. Very recently, they have been used for the particular problem we consider  in this paper of multi-view person image generation from single-view images~\cite{ma2017pose,zhao2017multiview}. While the results shown by both these approaches are  very  promising, they suffer from the same fundamental limitation in that   are methods trained in a fully supervised manner, that is, they   need to be trained with pairs of images of the same person dressing exactly the same clothes and under two different poses. This requires from specific datasets,  typically in the fashion domain~\cite{liu2016deepfashion,zheng2015scalable}. Tackling the problem in an unsupervised manner, one could leverage to an unlimited amount of  images and use other datasets for which no multi-view images of people are available.

In this paper we  therefore move a step  forward by  proposing a fully unsupervised GAN framework that, given a photo of a person, automatically generates  images of that person under new camera views and distinct body postures. The generative model we build is able to synthesize novel views of the body parts and clothes that are visible in the original image and also hallucinating those that are not seen. As shown in Fig.~\ref{fig:intro},  the generated images retain the body shape, and the new textures are consistent with the original image, even when input and desired poses are radically different.  In order to learn  this model using unlabeled data (i.e., our training data consists  of single images of people plus the  input and desired poses), we propose a GAN architecture that combines ingredients of the pose conditional adversarial networks~\cite{reed2016learning}, Cycle-GANs~\cite{zhu2017unpaired} and the loss functions used in image style transfer that aim at producing new images of high perceptual quality~\cite{gatys2016image}. 

More specifically, to circumvent the need  for pairs of training images of the same person under different poses, we split the problem in two main stages. First, we consider a pose conditioned bidirectional adversarial architecture which, given a single training photo, initially renders a new image under the desired pose. This synthesized image is then rendered-back to the original pose, hence being directly comparable to the input image. Second, in order to assess the quality of the rendered images we devise a  novel loss function computed over the $3$-tuple of images --original, rendered in the desired pose, and back-rendered to the original pose-- that incorporates  content and style terms. This function is conditioned  on the pose parameters and enforces the rendered image to retain the global semantic content of the original image as well as  its style at the joints location.

Extensive evaluation on the DeepFashion dataset~\cite{liu2016deepfashion} using unlabeled data shows very promising  results, even comparable with recent approaches trained in a fully supervised manner~\cite{ma2017pose,zhao2017multiview}.

\section{Related Work}
Rendering a person in an arbitrary pose from a single image  is a severely ill-posed problem as there are many cloth and body shape ambiguities caused by the new camera  view and the changing body pose, as well as large areas of missing data due to  body self-occlusions. Solving such a rendering problem requires thus introducing several sources of prior knowledge including, among others, the body shape, kinematic constraints, hair dynamics, cloth texture, reflectance models and fashion patterns.

Initial solutions to tackle this problem first built a 3D model  of the object and then synthesized the target images under the desired views~\cite{chen2013sweep,kholgade2014object,zheng2012interactive}. These methods, however, were constrained to rigid objects  defined by either CAD models or relatively simple geometric primitives.

More recently, with the advent of deep learning, there has been a growing interest in learning generative image models from data.  Several advanced models have been proposed for this purpose.  These include the variational autoencoders~\cite{kingma2013auto,lassner2017generative,rezende2014stochastic}, the autoregressive models~\cite{oord2016conditional,oord2016pixel}, and, most importantly the Generative Adversarial Networks~\cite{goodfellow2014generative}. 

GANs are very powerful generative models based on game theory. They   simultaneously train a generator network that produces synthetic samples (rendered images in our context) and a discriminator network that is trained to distinguish  between the generator's output and the true data. This idea is embedded by the so-called {\em adversarial loss}, which we shall use   in this paper to train our model.  GANs have been shown to produce very realistic images with a high level of detail.  They have been successfully used to render faces~\cite{karras2017progressive,radford2015unsupervised}, indoor scenes~\cite{karras2017progressive,wang2016generative} and clothes~\cite{iccv2017fashiongan}.

Particularly interesting for this work are those approaches that incorporate conditions to train GANs and constrain the generation process. Several conditions have been explored so far, such as discrete labels~\cite{mirza2014conditional,odena2016conditional}, and text~\cite{reed2016generative}. Images have also been used as a condition, for instance in the problem of image-to-image translation~\cite{isola2016image}, for future frame prediction~\cite{mathieu2016deep}, image inpainting~\cite{pathak2016context} and face alignment~\cite{huang2017beyond}. Very recently~\cite{iccv2017fashiongan} used both textual descriptions  and images as a condition to generate new clothing outfits. The works that are most related to ours  are~\cite{ma2017pose,zhao2017multiview}. They both propose GANs models for  the muti-view person image generation problem. However, the two approaches use  ground-truth supervision during train, i.e., pairs of images of the same person in two different poses dressed the same. Tackling the problem in a fully unsupervised manner, as we do in this paper, becomes a much harder task that requires more elaborate network designs, specially when estimating the loss of the rendered images. 

The unsupervised strategy we propose is somewhat related to that used in the Cycle-GANs~\cite{liu2017unsupervised,liu2016coupled,zhu2017unpaired} for image-to-image translation, also trained in the absence of paired examples. However, these approaches aim at   estimating a mapping between two distributions of images and no spatial transformation of the pixels in the input image are considered. This makes that the overall strategies and network architectures to address the two problems (image-to-image translation and multi-view generation) are essentially different.

\begin{figure*}[t!]
	\centering
	\includegraphics[width=\textwidth]{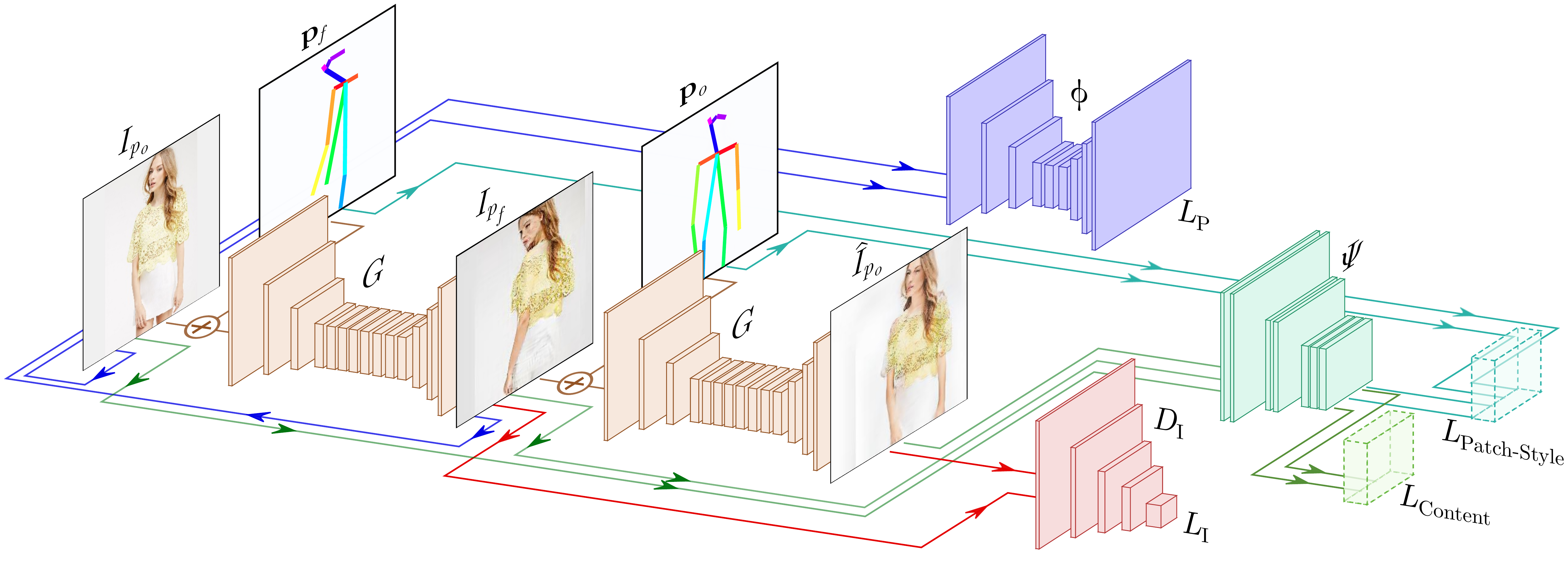}
	\caption{{\bf Overview of our unsupervised approach to generate multi-view   images of persons.} The proposed architecture consists of four main components: a generator $G$, a discriminator $D$, a 2D pose regressor $\Phi$ and the pre-trained feature extractor $\Psi$. Neither ground truth image nor any type of label is considered.}
	\label{fig:network}
\end{figure*}

\section{Problem Formulation}

Given a single-view image of a person, our goal is to train a GAN model in an unsupervised manner, allowing to generate photo-realistic pose transformations of the input image while retaining the person identity and clothes appearance. Formally, we seek to learn the mapping $(I_{p_o},\bp_f) \rightarrow I_{p_f}$ between an image $I_{p_o} \in \mathbb{R}^{3 \times H \times W}$ of a person with pose $\bp_o$ and the image $I_{p_f} \in \mathbb{R}^{3 \times H \times W}$ of the same person with the desired position $\bp_f$. Poses are represented by 2D skeletons with $N=18$  joints  $\bp=(\bu_1,\ldots,\bu_{N})$, where $\bu_i=(u_i,v_i)$ is the \textit{i}-th joint pixel location in the image. The model is trained in an unsupervised manner with training samples $\{ I_{p_o}^i,\bp_o^i,\bp_f^i\}_{i=1}^N$ that \textit{do not} contain the  ground-truth output image $I_{p_f}$.

\section{Method}
Figure~\ref{fig:network} shows an overview of our model. It is composed of four main modules: (1) A generator $G(I|\bp)$ that acts as a differentiable render    mapping one input image of a given person under a specific pose to an output image  of the same person under a different pose. Note that $G$ is used twice in our network, first to map the input image $I_{p_o}\rightarrow I_{p_f}$ and then to render the latter back to the original pose  $I_{p_f}\rightarrow \hat{I}_{p_o}$;  (2) A regressor $\Phi$   responsible of estimating the 2D joint locations of a given image; (3)  A discriminator  $D_{\text{I}}(I)$ that seeks to discriminate between generated and real samples; (4) A loss function, computed without ground truth, that aims to preserve the person identity. For this purpose, we devise a novel loss function that enforces   semantic content similarity of $I_{p_o}$ and $\hat{I}_{p_o}$, and style similarity between    $I_{p_o}$  and   $I_{p_f}$. 

In the following subsections we describe in detail each of these components as well as the 2D pose embedding we consider.

\subsection{Pose Embedding}

Drawing inspiration on~\cite{wei2016cpm}, the 2D location of each skeleton joint $\bu_i$ in an image $I\in \mathbb{R}^{3 \times H\times W}$ is represented as a probability density map $\bB_i\in \mathbb{R}^{H\times W}$ computed over the entire image domain as:
\begin{equation}
\bB_i [u,v]= P(\bu_i = (u,v))\;\; \forall \; (u,v)\in\mathcal{U} 
\end{equation}
being $\mathcal{U}$ the set of all $(u,v)$ pixel locations in the input image $I$. For each vertex $\bu_i$ we introduce a Gaussian peak with variance 0.03 in the position $(u_i,v_i)$ of the belief map  $\bB_i$. The full person pose $\bp$ is represented as the concatenation of all belief maps $\bp=(\bB_1,\ldots,\bB_{N})\in\mathbb{R}^{N \times H\times W}$.

\subsection{Network Architecture}

\paragraph{Generator.} 
Given an input image $I$ of a person, the generator $G(I|\bp)$ aims to render a photo-realistic image of that person in a desired pose $\bp$. In order to condition the generator with the pose we consider the concatenation $(I,\bp)\in\mathbb{R}^{ (N+3) \times H \times W}$ and feed this into a feed forward network that produces an output image  with the same dimensions as $I$. The generator is implemented as the variation of the network from Johnson~\etal~\cite{johnson2016perceptual} proposed by~\cite{zhu2017unpaired} as it achieved impressive results for the image-to-image translation problem.

\paragraph{Image Discriminator.} 
We implement the discriminator $D_{\text{I}}(I)$ as a PatchGan~\cite{isola2016image} network mapping  from the input image $I$ to a matrix  $\bY_{\text{I}}\in\mathbb{R}^{26 \times 26}$, where $\bY_{\text{I}}[i,j]$  represents the probability of the overlapping patch $ij$ to be real. This discriminator contains less parameters than other conventional discriminators typically used for GANs and enforces high frequency correctness to reduce the blurriness of the generated images.

\paragraph{Pose Detector.} Given an image $I$ of a person, $\Phi(I)$ is a 2D detection network responsible for estimating the skeleton joint locations $\bp \in \mathbb{R}^{N \times H\times W}$ in the image plane.  $\Phi(I)$ is implemented with the ResNet~\cite{he2016identity} based network by Zhu~\etal~\cite{zhu2017unpaired}.

\subsection{Learning the Model}

The loss function we define contains   three terms, namely an \textit{image adversarial loss}~\cite{goodfellow2014generative}  that pushes the distribution of the generated images to the distribution of the training images, the \textit{conditional pose loss } that enforces the pose of the generated images to be similar to the desired ones, and the  \textit{identity loss} that favors to preserve the person identity. We next describe each of these terms.

\paragraph{Image Adversarial Loss.}
In order to optimize the generator $G$ parameters and learn the distribution of the training  data,  we perform a standard \textit{min-max strategy game} between the generator and the image discriminator $D_{\text{I}}$. The generator and discriminator are jointly trained with the objective function $\mathcal{L}_{\text{I}}(G, D_{\text{I}}, I, \bp)$ where $D_{\text{I}}$ tries to maximize the probability of correctly classifying real and rendered images while $G$ tries to foul the discriminator. Formally, this loss is defined as:
 \begin{equation}
 \begin{split}
\mathcal{L}_{\text{I}}(G, D_{\text{I}}, I, \bp)& = \mathbb{E}_{I \sim p_{\text{data}}(I)}[\log D_{\text{I}}(I)]\\
&+ \mathbb{E}_{I \sim p_{\text{data}}(I)}[\log (1-D_{\text{I}}(G(I|\bp)))]
 \end{split}
 \end{equation}
 
 \paragraph{Conditional Pose Loss.}
 While reducing the \textit{image adversial loss}, the generator must also reduce the error produced by the 2D pose regressor $\Phi$. In this way, the generator not only learns to produce realistic samples but also learns how to generate samples consistent with the desired pose $\bp$. This loss is defined by:

 \begin{equation}
 \begin{split}
\mathcal{L}_{\text{P}}(G, \Phi, I, \bp)& = \| \Phi(G(I|\bp)) - \bp \|^2_2
\end{split}
 \end{equation}
 
 \paragraph{Identity Loss.}
 With the two previously defined losses $\mathcal{L}_{\text{I}}$ and $\mathcal{L}_{\text{P}}$ the generator is enforced to  generate realistic images of people in a desired position. However,  without  ground-truth supervision there is no constraint to guarantee  that the person identity --e.g., body shape, hair style -- in the original and rendered images is the same. In order to preserve person identity, we draw inspiration on the {\em content-style loss} that was previously introduced in~\cite{gatys2016image} to maintain high perceptual quality in  the problem of image style transfer. This loss consists of two main components, one to retain semantic similarity (`content') and the other to retain texture similarity (`style'). Based on this idea we define two sub-losses that aim at retaining the identity between the input image $\bI_{p_o}$ and the rendered image $\bI_{p_f}$. 
 
 For the {\em content} term, we argue that  the generator should be able to render-back the original image $I_{p_o}$ given the generated image $I_{p_f}$ and the original pose $\bp_o$, that is, $\hat{I}_o \approx I_{p_o}$, where    $\hat{I}_o=G(G(I_{p_o}|\bp_f)|\bp_o)$.   Nevertheless,  even when using PatchGan based discriminators, directly comparing  $I_{p_o}$ and $\hat{I}_{p_o}$ at a pixel level would struggle to handle high-frequency details leading to overly-smoothed images. Instead, we compare them based on their semantic content. Formally, we define the content loss to be:
 \begin{equation}
 \mathcal{L}_{\text{Content}} = \| \Psi_z(I_{p_o}) - \Psi_z(\hat{I}_{p_o}) \|^2_2
 \end{equation}
 where $\Psi_z(\cdot)$ represents the activations at the \textit{z}-th layer of a pretrained network. 	
 	
 In order to retain the {\em style}  of the original image into the rendered ones  we enforce the  texture around the visible joints of   $I_{p_o}$ and $I_{p_f}$ to be similar. This involves a  first step of extracting  -- in a differential manner --  patches  of  features around the   joints of   $I_{p_o}$ and $I_{p_f}$. More specifically, let $\Psi_z(I_{p_o})\in \mathbb{R}^{C \times H' \times W'}$ be the semantic features of $I_{p_o}$, and $\bB_{p_o} \in \mathbb{R}^{N \times H' \times W'}$ the down-sampled (using average pooling) probability maps associated to the pose $\bp_o$. The pose-conditioned patches are computed as:
  \begin{equation}
  \bX_{p_o,i} =\bB_{p_o,i}\cdot \Psi_z(\bI_{p_o}) \quad \forall i \in\{1,\ldots,N\}
  \end{equation} 
The representation of a patch style is then captured by the correlation between the different channels of its hidden representations $\bX_{p_o,i}$ using the spatial extend of the feature maps as the expectation. As previously done in~\cite{gatys2016image} this can be implemented by computing the   Gram matrix  $\mathcal{G}_{p_o,i} \in \mathbb{R}^{C \times C}$ for each patch $i$, defined  as the inner product between the vectorized feature maps of $\bX_{p_o,i}$.  The {\em Patch-Style} loss is then computed as the   mean square error between visible pairs of Gram matrices of the same joint in both images  $I_{p_o}$ and $I_{p_f}$:
\begin{equation}
\mathcal{L}_{\text{Patch-Style}}= \frac{1}{N} \sum_i^N \left ( \frac{\mathcal{G}_{p_o,i} -  \mathcal{G}_{p_f,i}}{H'W'} \right )^2
\label{eq:patchstyle}
\end{equation} 
  
Finally, we define the identity loss    as the weighted sum of the content and style losses:  
 \begin{equation}
 \begin{split}
 \mathcal{L}_{\text{Id}} &= \mathcal{L}_{\text{Content}}(\Psi, I_{p_o}, \hat{I}_{p_o}) \\
 &+ \lambda \mathcal{L}_{\text{Patch-Style}}(\Psi,  I_{p_o},  I_{p_f}, \bp_o, \bp_f)
 \end{split}
 \end{equation}
 where he parameter $\lambda$ controls the relative importance or the two    components.

 \paragraph{Full Loss.}
 We take the full loss as a linear combination of all previous loss terms:
   \begin{equation}
 \begin{split}
\mathcal{L}& = \mathcal{L}_{\text{I}}(G, D_{\text{I}}, I_{p_o}, \bp_f) + \lambda_{\text{P}} \mathcal{L}_{\text{P}}(G, \Phi, I_{p_o}, \bp_f)\\
&+  \mathcal{L}_{\text{I}}(G, D_{\text{I}}, I_{p_f}, \bp_o) +  \lambda_{\text{P}} \mathcal{L}_{\text{P}}(G, \Phi, I_{p_f}, \bp_o) \\
&+ \lambda_{\text{Id}} \mathcal{L}_{\text{Id}} + \lambda_{\text{P}} \mathcal{L}_{\Phi}(I, \bp_o)
\end{split}
 \end{equation}
where $\mathcal{L}_{\Phi}(I, \bp_o) = \| \Phi(I_{p_o}) -\bp_o \|^2_2$ is used to train the pose regressor $\Phi$. Our ultimate goal is to solve:
\begin{equation}
G^\star =\arg \min_G \max_{D_{\text{I}}, \Phi} \mathcal{L}
 \end{equation} 
  
 Some could argue that the terms $\mathcal{L}_{\text{I}}$ and $\mathcal{L}_{\text{P}}$  for the recovered image $\hat{I}_{p_o}$ are not required because the same information is expressed by $\mathcal{L}_{\text{Content}}$. However, we experienced that these two terms improved robustness and convergence properties during training.

\section{Implementation Details}
In order to  reduce the  model oscillation and obtain more photo-realistic results
we use the learning trick introduced in~\cite{mao2016multi} and replace the negative log likelihood of the \textit{adversarial loss} by  a least square loss. The image features $\Psi_z(I)$ are obtained from a pretrained VGG16~\cite{simonyan2014very} with $z=7$. We use Adam solver~\cite{kingma2014adam} with learning rate of 0.0002 for the generator, 0.0001 for the discriminators and a batch size 12. We train for 300 epochs with a linear decreasing rate after epoch 100.  The weights for the loss terms are set to $\lambda_{\text{P}}=700$ and  $\lambda_{\text{Id}}=0.3$. As in~\cite{shrivastava2016learning}, to improve training stability,  we update the discriminators using a buffer with the previous rendered images rather than those generated in the current iteration.  During training, the $\bp_f$ poses are randomly sampled from those in the training set.

\section{Experimental Evaluation}

We verify the effectiveness of our unsupervised GAN model through quantitative and qualitative evaluations. We next describe the dataset we used for evaluation and the results we obtained.  Supplementary material can be found on  \url{http://www.albertpumarola.com/research/person-synthesis/}.

\vspace{1mm}
\noindent{\bf Benchmark.} We have evaluated our approach on the  publicly available {\em In-shop Clothes Retrieval Benchmark} of the DeepFashion dataset~\cite{liu2016deepfashion}, that contains a large number of clothing images with diverse person poses.  Images of the dataset were initially resized to a fixed size of $256 \times 256$. We then applied data augmentation with all three possible flips per each image. After that, 2D pose was computed in all images using  the Convolutional Pose Machine (CPM)~\cite{wei2016cpm}, and  images for which CPM failed were removed from the dataset.  From the remaining images, we randomly selected 24,145 for training and 5,000 for test.  Test samples are also associated to a desired pose and its corresponding ground truth image, that will be used for quantitative evaluation purposes. Training images are only associated to a desired 2D pose. No ground truth warped image is considered during training.

\subsection{Quantitative results}
 Since test samples  are annotated with  ground truth images under  the desired pose, we can quantitatively evaluate the quality of the synthesis. Specifically, we use the metrics  considered by previous approaches on multi-view person generation~\cite{ma2017pose,zhao2017multiview}, namely the Structural Similarity (SSIM)~\cite{Wang2004ssim} and the Inception Score (IS)~\cite{salimans2016improved}. These are fairly standard metrics that focus more on the overall quality of the generated image rather than on  the pixel-level similarity between the generated image and the ground truth. Concretely, SSIM  models the changes in the structural information and IS give high scores for images with a large semantic content. 

  \begin{table}[t!]
  	\setlength{\tabcolsep}{3pt} 
  	\setlength\arrayrulewidth{0.9pt}
  	\centering
  	\resizebox{6.5cm}{!} {
  		\begin{tabular}{|l |c|c|}
  			\hline
  			\multicolumn{1}{|l|}{\cellcolor[gray]{0.95} \bf Method} & \multicolumn{1}{c|}{\cellcolor[gray]{0.95} \bf SSIM}   & \multicolumn{1}{c|}{\cellcolor[gray]{0.95} \bf IS} \\ 	
  			\hline\hline
  			Our Approach &  0.747   &  2.97	 \\
  			Ma \etal NIPS'2017~\cite{ma2017pose}&  0.762  & 3.09	 \\
  			Zhao \etal ArXiv'2017~\cite{zhao2017multiview} &  0.620  &  3.03	 \\
  			Sohn \etal NIPS'2015~\cite{Sohn2015learning}* &  0.580  &  2.35	 \\
  			Mirza \etal ArXiv'2014~\cite{mirza2014conditional}* &  0.590  &  2.45	 \\ 			
  			\hline
  		\end{tabular}}
  		\vspace{1.0mm}
  		\caption{\textbf{Quantitative Evaluation on the DeepFashion dataset.} SSIM and IS for our {\em unsupervised} approach and four {\em supervised} state-of-the-art methods. For all measures, the higher is better. `*' indicates that these results were taken from~\cite{zhao2017multiview}. Note: These results are just indicative, as the test splits in previous approaches are not available and may differ between the different methods of the table. Nevertheless, note that the quantitative results put our unsupervised approach on a par with other supervised approaches.}
        \vspace{-4mm}
  		\label{tab:quantitative}
  	\end{table}
 
In Table~\ref{tab:quantitative}  we report these scores for our approach and the two fully supervised methods~\cite{ma2017pose} and ~\cite{zhao2017multiview}, when evaluated on the DeepFashion~\cite{liu2016deepfashion} dataset. Two additional implementations of a Variational AutoEncoder (VAE)~\cite{Sohn2015learning} and a Conditional GAN (CGAN) model~\cite{mirza2014conditional}, reported in~\cite{zhao2017multiview}, are included.   It is worth to point that while all methods are evaluated on the same dataset, the test splits in each case are not the same. Therefore, the results on this table should be considered only as indicative. In any event, note that the two metrics indicate that the quality of the synthesis obtained by our unsupervised approach are very similar to the most recent supervised approaches and even outperform previous  VAE and CGAN implementations.

\begin{figure*}[t!]
	\centering
	\includegraphics[width=\textwidth]{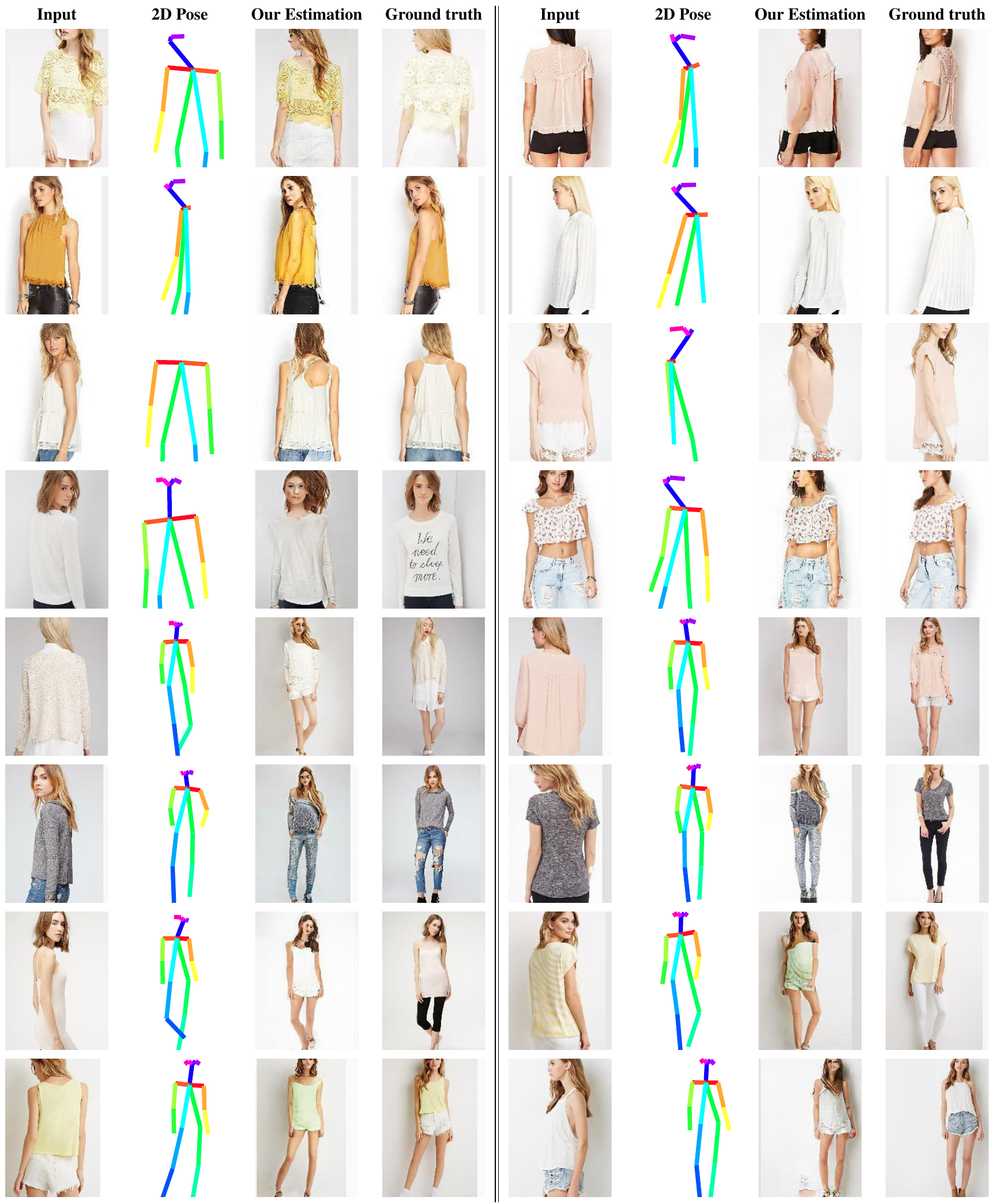}
	\caption{{\bf Test results on the DeepFashion~\cite{liu2016deepfashion} dataset.} Each test sample is represented by 4 images: input image, 2D desired pose, synthesized image and ground truth.}
	\label{fig:good_results}
\end{figure*}

\begin{figure*}[t!]
	\centering
	\includegraphics[width=\textwidth]{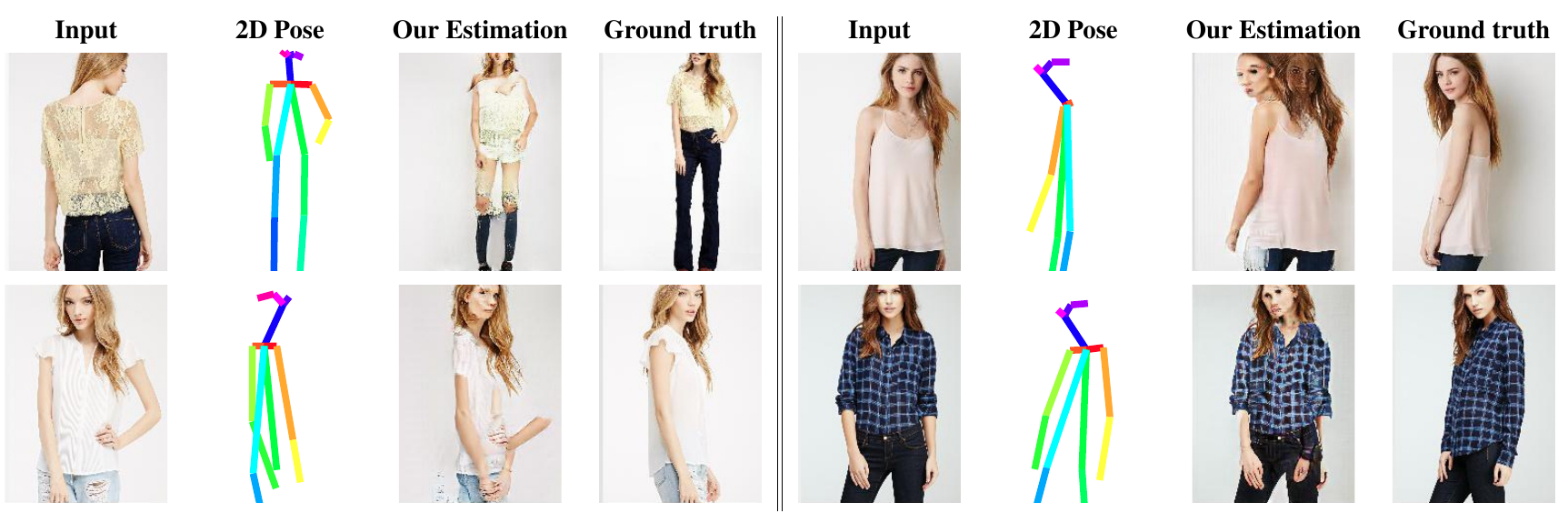}
	\caption{{\bf Test failures on the DeepFashion~\cite{liu2016deepfashion} dataset.}  We represent four different types of errors that typically occur in the failure cases (see text for details). }
	\label{fig:failures}
\end{figure*}
 
 \subsection{Qualitative results}
 We next present and discuss a series of qualitative results that will highlight the main characteristics of the proposed approach, including its ability to generalize to 
novel poses, to hallucinate image patches not observed in the original image and to render textures with high-frequency details. 
 
 In the Teaser image~\ref{fig:intro}  we observe all these characteristics. First, note the   ability of our GAN model to generalize to desired poses very different from that in the original image. In this case given a frontal image of the upper body of a woman, we show some of the generated images in which her pose is rotated by 180 $\deg$. In the right-most image of this example, the network is also able to hallucinate the two legs, not seen in the original image (despite not rendering the skirt). For this particular example, the network convincingly   renders  the high frequency details of the blouse. This is a very important characteristic of our model, and is a direct consequence of the loss function we have designed, and in particular of the term $\mathcal{L}_{\text{Patch-Style}}$ in Eq.~\eqref{eq:patchstyle} that aims at retaining the texture details of the original image into the generated one. This is in contrast to most of the renders generated by other GAN models~\cite{ma2017pose,zhao2017multiview,iccv2017fashiongan}, which typically wash out texture details. 
 
 Figure~\ref{fig:good_results}  presents another series of results obtained with our model.  In this case, each synthetically generated image is accompanied by the ground truth. Note again, the number of complex examples that are successfully addressed. Several cases show the hallucination of frontal poses from original poses facing back (or vice versa). Also are worth to mention those examples where the original image is in a side position with only one arm being observed, and the desired pose is either frontal of backwards, having to hallucinate both arms. Some of the textures of the t-shirts have very high frequency patterns and textures (example 4-th row/2-nd column, examples 6-th row) that are convincingly rendered under new poses.

 \vspace{1mm}
 \noindent{\bf Failure cases.} Tackling such an unconstrained problem in a fully unsupervised manner  causes a number of errors. We have roughly split them into four categories which we summarize in Figure~\ref{fig:failures}. The first type of error (top-left) is produced when textures in the original image are not correctly mapped onto the generated image. In this case, the partially observed dark trousers are transferred to a lower leg, resembling boots. In the top-right example, the face of the original image is not fully wash out in the new generated image. In the bottom-left we show a type of error which we denote as `geometric error', where the pose of the original image is not properly transferred to the target image. The bottom-right image  shows an example in which a part of the body in the original image (hand) is mapped as a texture in the synthesized one.

\begin{figure}[t!]
\begin{center}
\includegraphics[width=\linewidth, height=3.8cm]{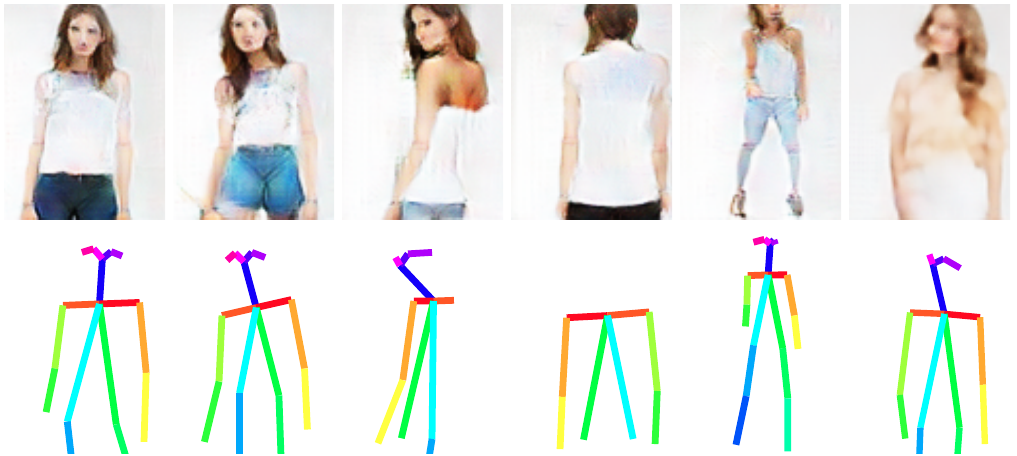}
\end{center}
 \caption{{\bf L1 vs Identity Loss.} Synthetic samples obtained by our model when it is trained with L1 loss and conditioned with the same inputs as in  Figure~\ref{fig:intro}. The first five columns correspond to $\hat{I}_{p_f}$, and the last column is the cycle image $\hat{I}_{p_o}$. Comparing these results  with those of Figure~\ref{fig:intro} it becomes clear that the L1 loss is not able to capture the person identity.}
\label{fig:l1}
\end{figure}

 \vspace{1mm}
 \noindent{\bf Ablation study.} Each component is crucial for the proper performance of the system. $D_{\text{I}}$ and $L_{\text{I}}$ constrain the system to generate realistic images; $\Phi$ and $L_{\text{P}}$ ensure the generator conditions the image generation to the given pose; and $\Psi$ and $L_{\text{Id}}$ force the generator to preserve the input image texture. Removing any of these elements would damage our network. For instance, Figure~\ref{fig:l1} shows the results when replacing $L_{\text{Id}}$ by the standard L1 loss used by most state-of-the-art GAN works. As it can observed in the last column of the figure, although $\hat{I}_{p_o}$ is preserving the low frequency texture of the original image, the person identity in $I_{p_f}$ is lost and all results tend to converge to a mean brunette woman with white t-shirt and blue jeans.

\begin{figure}[t!]
\begin{center}
\includegraphics[width=\linewidth, height=3.8cm]{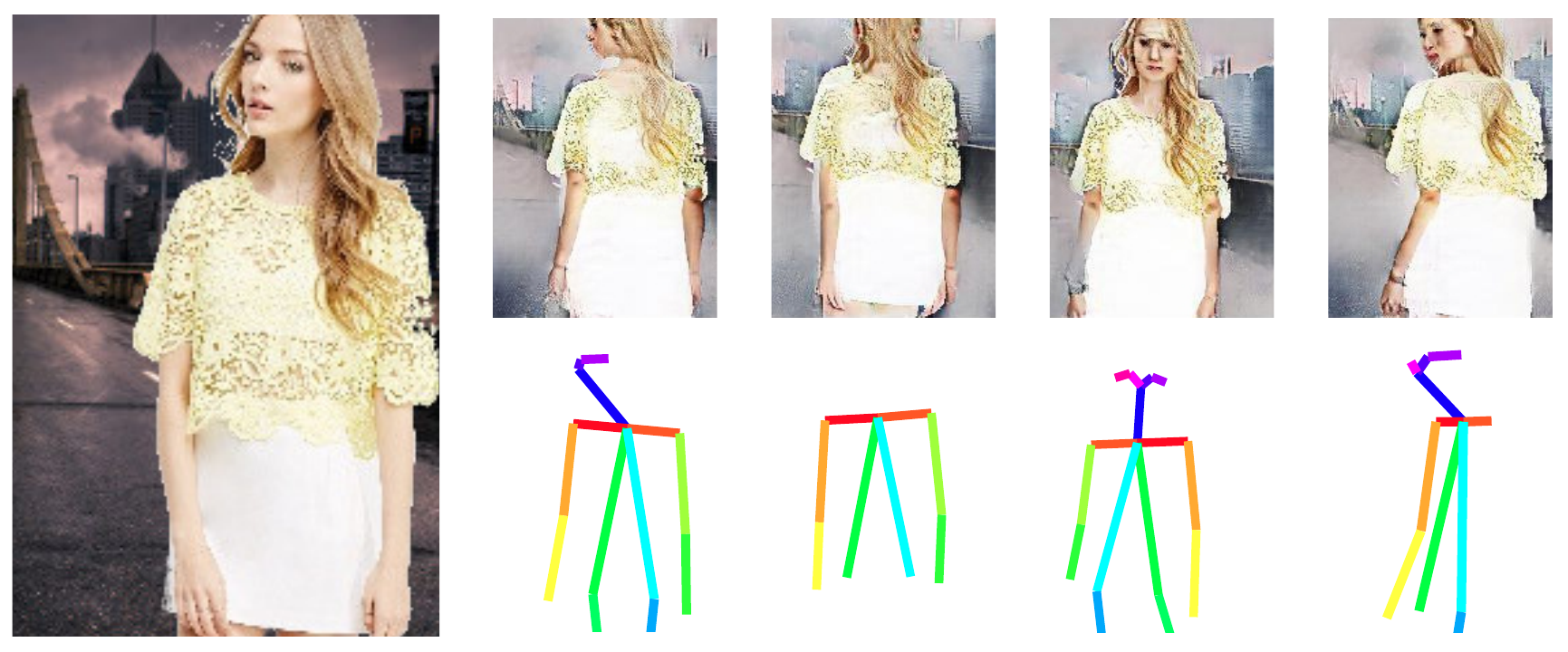}
\vspace{-5mm}
\end{center}
 \caption{{\bf Testing on images with background.} Given the original image of a person with background on the left and a desired body pose defined by a 2D skeleton (bottom-row), the model generates the person under that pose shown in the top-row. Albeit our model is trained with images with no background it does generalize fairly well to this situation (compare with the results of Figure~\ref{fig:intro}).}
\label{fig:background}
\end{figure}

 \vspace{1mm}
 \noindent{\bf Images with background.} To further test the limits of our model  Figure~\ref{fig:background} presents an evaluation of the model performance when the input image contains background. Surprisingly, although the model has no loss on background consistency nor was trained with images with background, the results are still very consistent. The person is  quite correctly rendered, while the background is over-smoothed. To become robust to background would require more complex datasets and specialized loss functions.

\section{Conclusion}
We have presented a novel approach for generating new images of a person under   arbitrary poses using a  GAN model that can be trained in a fully unsupervised manner.  This advances state-of-the-art, which so far, had only addressed the problem using supervision. To tackle this challenge, we have proposed an new framework that circumvents the need of training data by optimizing a loss function that only depends  on the input image and the rendered one, and aims at retaining the style and semantic content of the original image. Quantitative and qualitative evaluation on the DeepFashion~\cite{liu2016deepfashion} dataset shows very promising results, even for new body poses that highly differ from the input one and require hallucinating large portions of the image.  In the  future, we plan to further exploit our approach in other datasets (not only of humans) in the wild for which supervision is not possible. An important issue that will need to be addressed in this case, is the influence of complex backgrounds, and how they interfere in the generation process.  Finally, in order to improve the failure cases we have discussed, we will explore novel object- and geometry-aware  loss functions.

\vspace{1mm} 
\noindent{\bf Acknowledgments:} This work is supported in part by a Google Faculty Research Award, by the Spanish Ministry of Science and Innovation under projects HuMoUR TIN2017-90086-R, ColRobTransp DPI2016-78957 and  Mar\'ia de Maeztu Seal of Excellence MDM-2016-0656; and by the EU project AEROARMS ICT-2014-1-644271. We also thank Nvidia for hardware donation under the GPU Grant Program. 

{\small
\bibliographystyle{ieee}
\bibliography{egbib}
}

\balance

\end{document}